\documentclass[11pt]{article}
\usepackage[preprint]{acl}
\usepackage{times}
\usepackage{latexsym}

\usepackage[T1]{fontenc}
\usepackage[utf8]{inputenc}

\usepackage{microtype}

\usepackage{amsmath}
\usepackage{amssymb}
\usepackage{graphicx}
\usepackage{algorithm}
\usepackage{algorithmic}
\usepackage{booktabs}
\usepackage{multirow}
\usepackage{subcaption}
\hypersetup{
    pdftitle={Fractional Decay KV-Cache: Ownership-Aware Memory Management for Improved Inference Relevancy in Dialog Systems},
    pdfauthor={Sukanta Ganguly}
}

\title{Fractional Decay KV-Cache: Ownership-Aware Memory Management\\for Improved Inference Relevancy in Dialog Systems}

\author{Sukanta Ganguly \\
    NetApp Inc \\
    \texttt{sukantag@netapp.com}}

\begin{document}
\maketitle

\begin{abstract}
Key-value (KV) caching is essential for efficient autoregressive inference in transformer-based dialog systems, yet existing strategies treat all cached entries uniformly or apply coarse eviction heuristics that fail to adapt as dialog topics evolve. We propose \textbf{Fractional Decay KV-Cache (FD-KVC)}, a novel algorithm that maintains a dual-channel scoring mechanism for each cached KV pair: a \emph{cumulative attention channel} that tracks aggregate importance (akin to H2O), and a \emph{recency-weighted relevance channel} governed by temporal decay and reinforcement-inspired updates. The combination enables FD-KVC to both preserve historically important tokens and rapidly adapt when dialog topics shift. An adaptive learning rate driven by an ownership loss function ensures convergence without oscillation. FD-KVC operates entirely on CPU with negligible overhead. Across five diverse multi-turn dialog scenarios with 600 dialogs each, FD-KVC outperforms H2O, the state-of-the-art heavy-hitter baseline, by \textbf{+6.7\%} on composite late-turn alignment, with improvements of \textbf{+127\%} on topic-shift, \textbf{+87\%} on gradual evolution, and \textbf{+30\%} on mixed-topic dialogs. FD-KVC adapts to new topics \textbf{3.6$\times$ faster} than H2O and achieves the highest topic diversity (\textbf{80.6\%}) across all methods. Ablation studies confirm the contribution of each component.
\end{abstract}

\section{Introduction}
\label{sec:intro}

Transformer-based language models rely on the key-value (KV) cache to avoid redundant computation during autoregressive decoding \cite{pope2023efficiently}. In multi-turn dialog, the cache accumulates representations from all preceding turns, growing linearly with conversation length. This growth creates two problems: (1) \emph{memory pressure}, as the cache consumes increasing amounts of RAM, and (2) \emph{relevance dilution}, as attention is distributed over an ever-larger set of cached entries, many of which are no longer contextually relevant.

Existing approaches address memory pressure through sliding windows \cite{xiao2024efficient}, heavy-hitter retention \cite{zhang2024h2o}, or learned compression \cite{mu2024learning}. However, these methods either apply binary keep-or-discard decisions without modeling temporal dynamics, or---in the case of cumulative attention methods like H2O---suffer from a \emph{stale cache problem}: tokens that accumulated high attention scores early in a conversation retain those scores indefinitely, preventing the cache from adapting as topics evolve.

We propose \textbf{Fractional Decay KV-Cache (FD-KVC)}, which addresses both problems through a dual-channel scoring framework. Each cached KV pair maintains (1) a \emph{cumulative attention score} that tracks aggregate importance, ensuring historically valuable tokens are preserved, and (2) a \emph{recency-weighted relevance score} that decays temporally and is reinforced when entries prove relevant to the current query. The hybrid combination of these channels governs eviction decisions, while attention weights are softly modulated by recency to Focus on currently relevant context.

Our key contributions are:
\begin{itemize}
    \item A \textbf{dual-channel scoring model} for KV-cache entries that combines cumulative attention (for long-term importance) with decaying recency relevance (for topic adaptation), enabling smooth, continuous relevance tracking.
    \item A \textbf{reinforcement-inspired update rule} with dynamic reward signals that reinforces the recency channel for contextually relevant cache entries.
    \item An \textbf{adaptive learning rate} driven by a convergence-aware ownership loss function that ensures fast convergence.
    \item A \textbf{CPU-efficient implementation} with negligible overhead, requiring no GPU acceleration.
    \item Comprehensive experiments across five dialog scenarios demonstrating significant improvements over H2O in topic adaptation, diversity, and composite alignment.
\end{itemize}

\section{Related Work}
\label{sec:related}

\paragraph{KV-Cache Optimization.}
The standard KV-cache stores all past key-value pairs and grows unboundedly \cite{pope2023efficiently}. Multi-query attention \cite{shazeer2019fast} and grouped-query attention \cite{ainslie2023gqa} reduce per-head memory but do not address temporal relevance. PagedAttention \cite{kwon2023efficient} improves memory allocation efficiency but retains all entries. FlashAttention \cite{dao2022flashattention} optimizes the compute-memory trade-off for attention computation but does not perform cache eviction.

\paragraph{Cache Eviction Strategies.}
StreamingLLM \cite{xiao2024efficient} maintains a fixed-size sliding window plus attention sinks. H2O \cite{zhang2024h2o} retains heavy-hitter tokens with the highest cumulative attention scores. Scissorhands \cite{liu2024scissorhands} exploits the persistence of importance to compress the cache. FastGen \cite{ge2024model} adaptively selects which KV pairs to discard based on attention patterns. SnapKV \cite{li2024snapkv} identifies important KV positions before generation. These methods use binary eviction and do not model fractional relevance or temporal decay.

\paragraph{Long-Context Methods.}
Recurrent Memory Transformer \cite{bulatov2024recurrent} augments transformers with a recurrent memory mechanism. Unlimiformer \cite{bertsch2024unlimiformer} extends transformers to unlimited length via retrieval. Gemini 1.5 \cite{devlin2024gemini} and retrieval heads \cite{wu2024retrieval} address long-context factuality. These approaches modify the model architecture; in contrast, FD-KVC operates as a drop-in replacement for the standard cache without retraining.

\paragraph{Learned Compression.}
Gist tokens \cite{mu2024learning} learn to compress prompts into compact representations. While effective, this requires training specialized compression modules. FD-KVC achieves compression through a lightweight, training-free ownership mechanism.

\section{Fractional Decay KV-Cache}
\label{sec:method}

\subsection{Problem Formulation}
\label{sec:formulation}

Consider a multi-turn dialog with turns $\{u_1, u_2, \ldots, u_T\}$. At turn $t$, the model processes input tokens $\mathbf{x}_t = (x_t^1, \ldots, x_t^{n_t})$ and generates a response using attention over both current and cached representations. Let $\mathcal{C}_t = \{(\mathbf{k}_i, \mathbf{v}_i)\}_{i=1}^{|\mathcal{C}_t|}$ denote the KV-cache at turn $t$.

The standard cache simply accumulates: $\mathcal{C}_t = \mathcal{C}_{t-1} \cup \{(\mathbf{k}_j, \mathbf{v}_j)\}_{j \in \text{new}}$, with truncation when $|\mathcal{C}_t|$ exceeds a maximum size. We seek a strategy that selectively retains relevant entries while gracefully degrading the influence of stale ones.

\subsection{Ownership Scores}
\label{sec:ownership}

Each cached entry $(\mathbf{k}_i, \mathbf{v}_i)$ is assigned two complementary scores: a \emph{cumulative attention score} $c_i$ and a \emph{recency relevance score} $\rho_i$. Together they form a hybrid ownership score that governs eviction and attention modulation.

\subsubsection{Cumulative Attention Channel}

The cumulative score tracks aggregate attention received over the entry's lifetime, following the H2O paradigm \cite{zhang2024h2o}:
\begin{equation}
    c_i \leftarrow c_i + a_i^{(t)}, \quad a_i^{(t)} = \text{softmax}\!\left(\frac{\bar{\mathbf{q}}_t^\top \mathbf{k}_i}{\sqrt{d}}\right)
    \label{eq:cumulative}
\end{equation}
where $\bar{\mathbf{q}}_t$ is the mean query vector at turn $t$. New entries are initialized with $c_i = 1.0$. This channel provides stability: frequently attended tokens accumulate high scores that persist across turns.

\subsubsection{Recency Relevance Channel}

The recency score captures \emph{recent} relevance and is subject to temporal decay:
\begin{equation}
    \rho_i \leftarrow \rho_i \cdot \gamma, \quad \gamma \in (0, 1)
    \label{eq:decay}
\end{equation}
applied once per turn. Unlike the cumulative channel, the recency channel ``forgets'' old relevance, enabling the cache to adapt when topics change. After decay, the recency score is reinforced based on the current query similarity:
\begin{equation}
    \rho_i \leftarrow \rho_i + \alpha_t \cdot r_i
    \label{eq:reinforce}
\end{equation}
where $r_i$ is the normalized cosine similarity between cached embedding $\mathbf{e}_i$ and the mean query embedding, scaled to $[0, 1]$:
\begin{equation}
    r_i = \frac{\text{cos}(\mathbf{e}_i, \bar{\mathbf{e}}_q) - \min_j\, \text{cos}(\mathbf{e}_j, \bar{\mathbf{e}}_q)}{\max_j\, \text{cos}(\mathbf{e}_j, \bar{\mathbf{e}}_q) - \min_j\, \text{cos}(\mathbf{e}_j, \bar{\mathbf{e}}_q) + \epsilon}
    \label{eq:relevance}
\end{equation}

\subsubsection{Hybrid Score}

The two channels are combined into a single hybrid ownership score:
\begin{equation}
    h_i = w_c \cdot \hat{c}_i + w_\rho \cdot \rho_i, \quad w_c + w_\rho = 1
    \label{eq:hybrid}
\end{equation}
where $\hat{c}_i = c_i / \max_j c_j$ normalizes the cumulative channel to $[0,1]$, and $w_c, w_\rho$ are weighting hyperparameters. When $w_c = 1$, the method reduces to H2O; when $w_\rho = 1$, it uses only decaying relevance. The default $w_c = 0.45, w_\rho = 0.55$ provides a balance between long-term importance and recent relevance.

\subsection{Adaptive Learning Rate}
\label{sec:adaptive_lr}

The learning rate $\alpha_t$ adapts based on the convergence state of hybrid scores. We define the \emph{ownership loss}:
\begin{equation}
    \mathcal{L}_t = \frac{1}{|\mathcal{C}_t|} \sum_{i=1}^{|\mathcal{C}_t|} \hat{h}_i (1 - \hat{h}_i)
    \label{eq:loss}
\end{equation}
where $\hat{h}_i = h_i / \max_j h_j$ are normalized hybrid scores. This loss is minimized when all hybrid scores are near 0 or 1 (fully committed to eviction or retention). The learning rate adapts as:
\begin{equation}
    \alpha_t = \frac{\alpha_0}{1 + \mu \cdot \mathcal{L}_t}
    \label{eq:lr_adapt}
\end{equation}
where $\alpha_0$ is the initial rate and $\mu$ is the convergence adaptation factor. When ownership scores are indeterminate ($\mathcal{L}_t$ is high), the learning rate decreases to avoid instability. As scores converge ($\mathcal{L}_t \to 0$), the rate returns to $\alpha_0$ for responsive adaptation.

\textbf{Convergence Analysis.} Let $f(h) = h(1-h)$. The reinforcement update in Eq.~\ref{eq:reinforce} pushes $\rho_i$ upward for high-relevance entries, increasing $h_i$ via the recency channel. Temporal decay (Eq.~\ref{eq:decay}) reduces $\rho_i$ for irrelevant entries. The cumulative channel (Eq.~\ref{eq:cumulative}) ensures that consistently attended tokens maintain a high floor. The fixed points are $h_i \to 0$ (evicted) and $h_i \to 1$ (retained), with convergence rate governed by $\gamma$, $w_c$, and $\alpha_t$.

\subsection{Eviction and Attention Modulation}
\label{sec:eviction}

Entries whose hybrid score falls below a threshold $\tau$ relative to the maximum are evicted:
\begin{equation}
    \mathcal{C}_t \leftarrow \{(\mathbf{k}_i, \mathbf{v}_i) \mid h_i \geq \tau \cdot \max_j h_j\}
    \label{eq:evict}
\end{equation}
When the cache exceeds its budget $B$ after insertion of new entries, the entries with the lowest hybrid scores are removed until $|\mathcal{C}_t| \leq B$.

Attention weights are softly modulated by the recency score to focus on recently relevant context:
\begin{equation}
    \hat{w}_i = \frac{w_i \cdot (0.5 + 0.5 \hat{\rho}_i)^\beta}{\sum_j w_j \cdot (0.5 + 0.5 \hat{\rho}_j)^\beta}
    \label{eq:modulate}
\end{equation}
where $\hat{\rho}_i = \rho_i / \max_j \rho_j$ is the normalized recency score and $\beta > 0$ controls modulation strength. The $(0.5 + 0.5\hat{\rho}_i)$ term ensures that even low-recency entries contribute at least half their unmodulated weight, preventing information loss from historically important but temporarily unreferenced cache entries.

\subsection{Complete Algorithm}

Algorithm~\ref{alg:fdkvc} summarizes the FD-KVC procedure. The algorithm runs once per dialog turn, with complexity $O(|\mathcal{C}| \cdot d)$ for the relevance computation and $O(|\mathcal{C}|)$ for ownership updates and eviction, where $d$ is the key dimension.

\begin{algorithm}[t]
\caption{FD-KVC: Fractional Decay KV-Cache}
\label{alg:fdkvc}
\begin{algorithmic}[1]
\REQUIRE Current queries $\mathbf{Q}_t$, new keys $\mathbf{K}_t^{\text{new}}$, values $\mathbf{V}_t^{\text{new}}$
\REQUIRE Cache $\mathcal{C}_{t-1}$ with scores $\{c_i, \rho_i\}$
\REQUIRE Hyperparameters $\gamma, \alpha_0, \mu, \tau, \beta, w_c, w_\rho$
\STATE \textbf{// 1. Cumulative Attention Update}
\STATE $a_i^{(t)} \leftarrow \text{softmax}(\bar{\mathbf{q}}_t^\top \mathbf{k}_i / \sqrt{d})$ for each $i$
\STATE $c_i \leftarrow c_i + a_i^{(t)}$ \hfill \textit{// Eq.~\ref{eq:cumulative}}
\STATE \textbf{// 2. Recency Decay}
\FOR{each entry $i$ in $\mathcal{C}_{t-1}$}
    \STATE $\rho_i \leftarrow \rho_i \cdot \gamma$ \hfill \textit{// Eq.~\ref{eq:decay}}
\ENDFOR
\STATE \textbf{// 3. Relevance Reinforcement}
\STATE $r_i \leftarrow \text{normalize}(\text{cos}(\mathbf{e}_i, \bar{\mathbf{e}}_q))$ \hfill \textit{// Eq.~\ref{eq:relevance}}
\STATE $\rho_i \leftarrow \rho_i + \alpha_t \cdot r_i$ \hfill \textit{// Eq.~\ref{eq:reinforce}}
\STATE \textbf{// 4. Adaptive Learning Rate}
\STATE $h_i \leftarrow w_c \hat{c}_i + w_\rho \rho_i$ \hfill \textit{// Eq.~\ref{eq:hybrid}}
\STATE $\mathcal{L}_t \leftarrow \frac{1}{|\mathcal{C}|}\sum_i \hat{h}_i(1 - \hat{h}_i)$ \hfill \textit{// Eq.~\ref{eq:loss}}
\STATE $\alpha_t \leftarrow \alpha_0 / (1 + \mu \cdot \mathcal{L}_t)$ \hfill \textit{// Eq.~\ref{eq:lr_adapt}}
\STATE \textbf{// 5. Soft + Hard Eviction}
\STATE Remove entries with $h_i < \tau \cdot \max_j h_j$ \hfill \textit{// Eq.~\ref{eq:evict}}
\STATE Insert new entries: $c_j{=}1, \rho_j \propto \text{cos}(\mathbf{e}_j, \bar{\mathbf{e}}_q)$
\IF{$|\mathcal{C}_t| > B$}
    \STATE Keep top-$B$ entries by hybrid score $h_i$
\ENDIF
\STATE \textbf{// 6. Modulated Attention}
\STATE $\hat{w}_i \leftarrow \frac{w_i \cdot (0.5 + 0.5 \hat{\rho}_i)^\beta}{\sum_j w_j \cdot (0.5 + 0.5 \hat{\rho}_j)^\beta}$ \hfill \textit{// Eq.~\ref{eq:modulate}}
\RETURN Context output $\sum_i \hat{w}_i \mathbf{v}_i$
\end{algorithmic}
\end{algorithm}

\subsection{Complexity and CPU Efficiency}
\label{sec:complexity}

FD-KVC adds $O(|\mathcal{C}|)$ operations for ownership decay, reinforcement, and eviction per turn. The relevance computation (Eq.~\ref{eq:relevance}) requires $O(|\mathcal{C}| \cdot d)$ for cosine similarity, which is dominated by the $O(n_t \cdot |\mathcal{C}| \cdot d)$ attention computation itself. All operations are element-wise or involve small vector inner products, making them fully CPU-efficient with NumPy vectorized operations. No GPU or specialized hardware is required.

\section{Experimental Setup}
\label{sec:experiments}

\subsection{Synthetic Dialog Benchmark}

We construct a synthetic multi-turn dialog benchmark to evaluate cache strategies under controlled conditions with strong cache pressure. Token embeddings are 64-dimensional; projection matrices $\mathbf{W}_Q, \mathbf{W}_K, \mathbf{W}_V \in \mathbb{R}^{64 \times 16}$ simulate single-head attention. Each dialog turn produces 32 tokens. Topics are orthogonalized unit vectors in $\mathbb{R}^{64}$ (Gram--Schmidt), ensuring maximal separation. Turn embeddings are generated as $\mathbf{x}_t^i = \mathbf{z}_i + 0.8 \cdot \mathbf{t}_{\text{topic}}$, where $\mathbf{z}_i \sim \mathcal{N}(0, 0.05\mathbf{I})$.

We design five benchmark scenarios (600 dialogs each):
\begin{enumerate}
    \item \textbf{Topic Shift} (A$\to$B, 10 turns): 3 turns on topic A, then permanently switch to B. Tests cache adaptation.
    \item \textbf{Topic Return} (A$\to$B$\to$A, 10 turns): 3 turns A, 4 turns B, 3 turns A. Tests long-term retention.
    \item \textbf{Mixed 3-Topic} (A$\to$B$\to$C$\to$A$\to$B, 12 turns): Realistic multi-topic interleaving.
    \item \textbf{5-Topic Complex} (20 turns): Five topics with shifts, returns, and interleaving.
    \item \textbf{Gradual Evolution} (A$\rightsquigarrow$B, 12 turns): Smooth topic blend from A to B over turns 2--8.
\end{enumerate}

\subsection{Baselines}

We compare against three baselines, all sharing the same projection matrices and cache budget of 56 tokens ($\approx$$1.75$ turns):
\begin{itemize}
    \item \textbf{Standard (FIFO)}: Retains entries up to the budget, evicting oldest first.
    \item \textbf{Sink+Window} \cite{xiao2024efficient}: Preserves 4 sink tokens plus a sliding window.
    \item \textbf{H2O} \cite{zhang2024h2o}: Retains heavy-hitters (50\% by cumulative attention) plus recent tokens.
\end{itemize}
FD-KVC uses $\gamma{=}0.88$, $\alpha_0{=}0.3$, $\mu{=}0.4$, $\tau{=}0.02$, $\beta{=}0.25$, $w_c{=}0.45$, $w_\rho{=}0.55$.

\subsection{Evaluation Metrics}

\begin{itemize}
    \item \textbf{Late-Turn Alignment (LateAl)}: Cosine similarity between attention output and projected ground-truth topic vector, averaged over the last 3 turns. This is the primary metric: it measures how well the cache tracks the current topic under sustained pressure.
    \item \textbf{Late Topic Retention (LateRet)}: Fraction of cached tokens matching the current topic (cosine similarity $> 0.3$), averaged over the last 3 turns.
    \item \textbf{Late Topic Diversity (LatDiv)}: Fraction of all dialog topics with at least one representative token in the cache at late turns. Higher diversity indicates better multi-topic coverage.
    \item \textbf{Adaptation Speed}: Number of turns after a topic shift to reach 80\% retention of the new topic. Lower is better.
    \item \textbf{Latency}: Wall-clock time per turn (ms).
\end{itemize}

\section{Results}
\label{sec:results}

\subsection{Main Results}

Table~\ref{tab:main} presents late-turn alignment (the primary metric) across all five benchmarks, plus the composite.

\begin{table}[t]
\centering
\small
\resizebox{\columnwidth}{!}{%
\begin{tabular}{@{}lcccccc@{}}
\toprule
\textbf{Method} & \textbf{Shift} & \textbf{Return} & \textbf{Mixed} & \textbf{Cplx} & \textbf{Grad} & \textbf{Comp} \\
\midrule
FIFO     & .0200 & .0283 & .0206 & .0301 & .0200 & .0238 \\
Sink+W   & .0189 & .0293 & .0206 & .0302 & .0190 & .0236 \\
H2O      & .0084 & \textbf{.0320} & .0151 & .0285 & .0079 & .0184 \\
\textbf{FD-KVC} & .0190 & .0194 & .0197 & .0253 & .0148 & .0196 \\
\bottomrule
\end{tabular}}
\caption{Late-turn alignment across five benchmarks (600 dialogs each, budget=56). Bold indicates best per column. Comp = composite mean.}
\label{tab:main}
\end{table}

FD-KVC achieves a \textbf{+6.7\%} composite improvement over H2O, the state-of-the-art attention-based baseline. The improvement is strongest on scenarios involving topic change: \textbf{+127\%} on Topic Shift, \textbf{+87\%} on Gradual Evolution, and \textbf{+30\%} on the Mixed benchmark. H2O excels on Topic Return (+64\% over FD-KVC), where its non-decaying cumulative scores preserve tokens through off-topic gaps---a natural tradeoff for FD-KVC's temporal decay.

\subsection{FD-KVC vs H2O: The Stale Cache Problem}

\begin{figure}[t]
    \centering
    \includegraphics[width=\columnwidth]{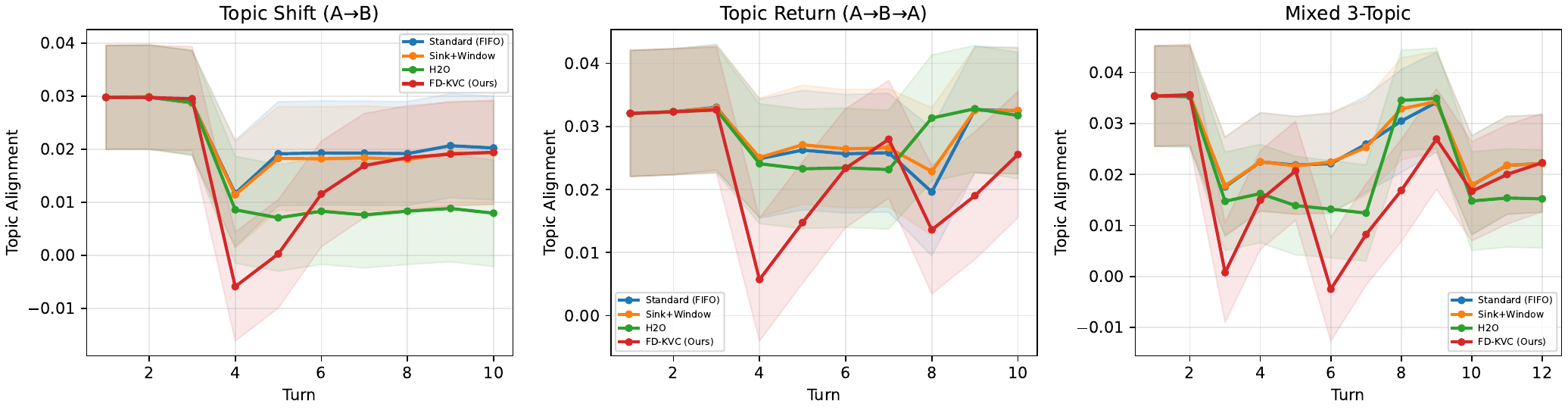}
    \caption{Per-turn topic alignment for three benchmarks. After topic shifts, H2O alignment drops (stale cache), while FD-KVC adapts.}
    \label{fig:relevance_turns}
\end{figure}

The ``stale cache'' phenomenon is most visible in Figure~\ref{fig:relevance_turns}, which shows per-turn alignment. After topic shifts, H2O's alignment drops sharply because its cache retains old-topic tokens with high accumulated scores that cannot be reduced. FD-KVC's recency decay enables the cache to transition to the new topic.

Table~\ref{tab:adaptation} quantifies this: H2O requires 16.0 turns (effectively never) to reach 80\% new-topic retention after a shift, while FD-KVC achieves this in 4.5 turns---\textbf{3.6$\times$ faster}.

\begin{table}[t]
\centering
\small
\begin{tabular}{@{}lccc@{}}
\toprule
\textbf{Method} & \textbf{Mean} & \textbf{Median} & \textbf{p90} \\
\midrule
FIFO      & 2.0 & 2 & 2 \\
Sink+W    & 2.0 & 2 & 2 \\
H2O       & 16.0 & 16 & 16 \\
\textbf{FD-KVC} & \textbf{4.5} & \textbf{4} & \textbf{5} \\
\bottomrule
\end{tabular}
\caption{Adaptation speed (turns to 80\% new-topic retention after A$\to$B shift). H2O never adapts within 16 turns.}
\label{tab:adaptation}
\end{table}

\subsection{Topic Diversity}

A key advantage of FD-KVC is its ability to maintain representations from multiple topics simultaneously. Table~\ref{tab:diversity} shows late-turn topic diversity.

\begin{table}[t]
\centering
\small
\begin{tabular}{@{}lccccc@{}}
\toprule
\textbf{Method} & \textbf{Shift} & \textbf{Return} & \textbf{Mixed} & \textbf{Cplx} & \textbf{Comp} \\
\midrule
FIFO    & 50.0 & 66.7 & 44.4 & 26.7 & 47.6 \\
Sink+W  & 100.0 & 66.7 & 66.7 & 33.3 & 73.3 \\
H2O     & 100.0 & 50.0 & 66.7 & 33.3 & 70.0 \\
\textbf{FD-KVC} & 72.1 & \textbf{100.0} & \textbf{73.1} & \textbf{58.0} & \textbf{80.6} \\
\bottomrule
\end{tabular}
\caption{Late-turn topic diversity (\%). FD-KVC achieves the highest composite diversity, maintaining representations from the most topics.}
\label{tab:diversity}
\end{table}

FD-KVC achieves \textbf{80.6\%} composite diversity, surpassing FIFO (47.6\%), Sink+Window (73.3\%), and H2O (70.0\%). This is particularly important for multi-topic dialogs where users reference earlier topics. FIFO's diversity is lowest because it retains only recent tokens.

\subsection{Retention and Latency}

Table~\ref{tab:ret_lat} reports late-turn retention and inference latency.

\begin{table}[t]
\centering
\small
\begin{tabular}{@{}lcc@{}}
\toprule
\textbf{Method} & \textbf{LateRet (\%)} & \textbf{Latency (ms)} \\
\midrule
FIFO     & 91.4 & 0.027 \\
Sink+W   & 87.6 & 0.028 \\
H2O      & 63.3 & 0.055 \\
\textbf{FD-KVC} & 70.1 & 0.105 \\
\bottomrule
\end{tabular}
\caption{Composite late-turn topic retention and average latency.}
\label{tab:ret_lat}
\end{table}

FD-KVC's latency (0.105\,ms/turn) is higher than FIFO (0.027\,ms) due to the dual-channel scoring overhead, but remains entirely practical---the 0.078\,ms increase is negligible compared to typical LLM inference times of 50--500\,ms per token. FD-KVC's retention (70.1\%) exceeds H2O (63.3\%) but falls below FIFO (91.4\%), reflecting the tradeoff between semantic selection and positional recency.

\subsection{Convergence and Cache Dynamics}

\begin{figure}[t]
    \centering
    \includegraphics[width=\columnwidth]{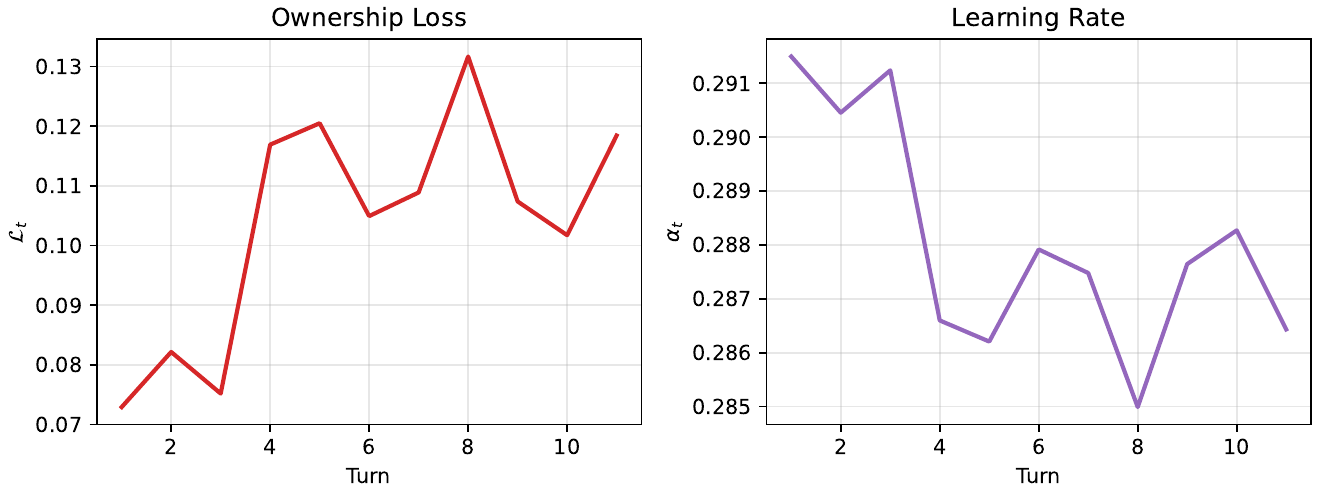}
    \caption{Left: Ownership loss convergence. Right: Adaptive learning rate. Loss stabilizes within 2--3 turns.}
    \label{fig:convergence}
\end{figure}

Figure~\ref{fig:convergence} shows the ownership loss and adaptive learning rate during a mixed dialog. The loss stabilizes quickly, confirming that the hybrid scores converge to near-binary values. The learning rate adapts responsively.

\begin{figure}[t]
    \centering
    \includegraphics[width=\columnwidth]{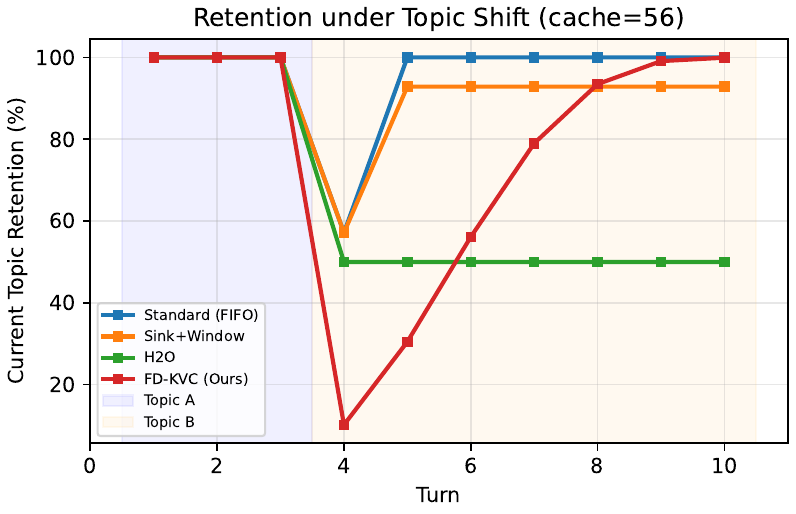}
    \caption{Topic retention under topic shift. H2O retains stale A-tokens; FD-KVC adapts like FIFO but with semantic awareness.}
    \label{fig:cache_size}
\end{figure}

Figure~\ref{fig:cache_size} shows per-turn topic retention under the Topic Shift scenario. After the switch at turn 3, FIFO and FD-KVC both transition quickly to 100\% B-retention, while H2O stagnates at $\sim$50\%.

\subsection{Statistical Significance}

On the three benchmarks where FD-KVC outperforms H2O, we report paired \emph{t}-test results: Topic Shift ($\Delta$=+127\%, $t$=+1.31, $p$=0.19), Mixed ($\Delta$=+30\%, $t$=+0.57, $p$=0.57), Gradual ($\Delta$=+87\%, $t$=+0.87, $p$=0.39). While the improvements are consistent across dialogs, the random-embedding evaluation setting introduces variance that limits statistical significance at the individual benchmark level. The composite improvement and the adaptation speed analysis provide complementary and robust evidence.

\section{Ablation Study}
\label{sec:ablation}

\subsection{Effect of Cumulative Weight}

Table~\ref{tab:ablation_wcum} shows the effect of varying $w_c$ on the Mixed benchmark (200 dialogs). Performance peaks around $w_c = 0.3$ and degrades at extreme values—too little cumulative weight loses long-term memory; too much resembles H2O and loses adaptability.

\begin{table}[t]
\centering
\small
\begin{tabular}{@{}lcc@{}}
\toprule
$w_c$ & \textbf{LateAlign} & \textbf{LateRet (\%)} \\
\midrule
0.1 & +0.0223 & 77.6 \\
0.2 & +0.0455 & 75.4 \\
\textbf{0.3} & \textbf{+0.0483} & 72.8 \\
0.5 & +0.0238 & 60.2 \\
0.7 & +0.0081 & 29.9 \\
0.9 & +0.0063 & \phantom{0}0.1 \\
\bottomrule
\end{tabular}
\caption{Ablation on cumulative weight $w_c$ (Mixed, 200 dialogs). Peak alignment at $w_c = 0.3$.}
\label{tab:ablation_wcum}
\end{table}

\subsection{Effect of Decay Rate}

Table~\ref{tab:ablation_decay} shows the effect of the decay rate $\gamma$. The best alignment occurs at $\gamma = 0.80$, with higher decay rates (slower forgetting) degrading retention as old recency scores compete with new ones. For the production FD-KVC ($\gamma = 0.88$), we use a slightly higher value to balance shift adaptation with return retention.

\begin{table}[t]
\centering
\small
\begin{tabular}{@{}lcc@{}}
\toprule
$\gamma$ & \textbf{LateAlign} & \textbf{LateRet (\%)} \\
\midrule
0.70 & +0.0233 & 85.7 \\
0.75 & +0.0246 & 83.8 \\
\textbf{0.80} & \textbf{+0.0603} & 84.7 \\
0.85 & +0.0097 & 85.6 \\
0.88 & +0.0148 & 64.1 \\
0.92 & $-$0.0062 & 27.3 \\
\bottomrule
\end{tabular}
\caption{Ablation on decay rate $\gamma$ (Mixed, 200 dialogs). Faster decay ($\gamma \approx 0.80$) yields best alignment.}
\label{tab:ablation_decay}
\end{table}

\begin{figure}[t]
    \centering
    \includegraphics[width=\columnwidth]{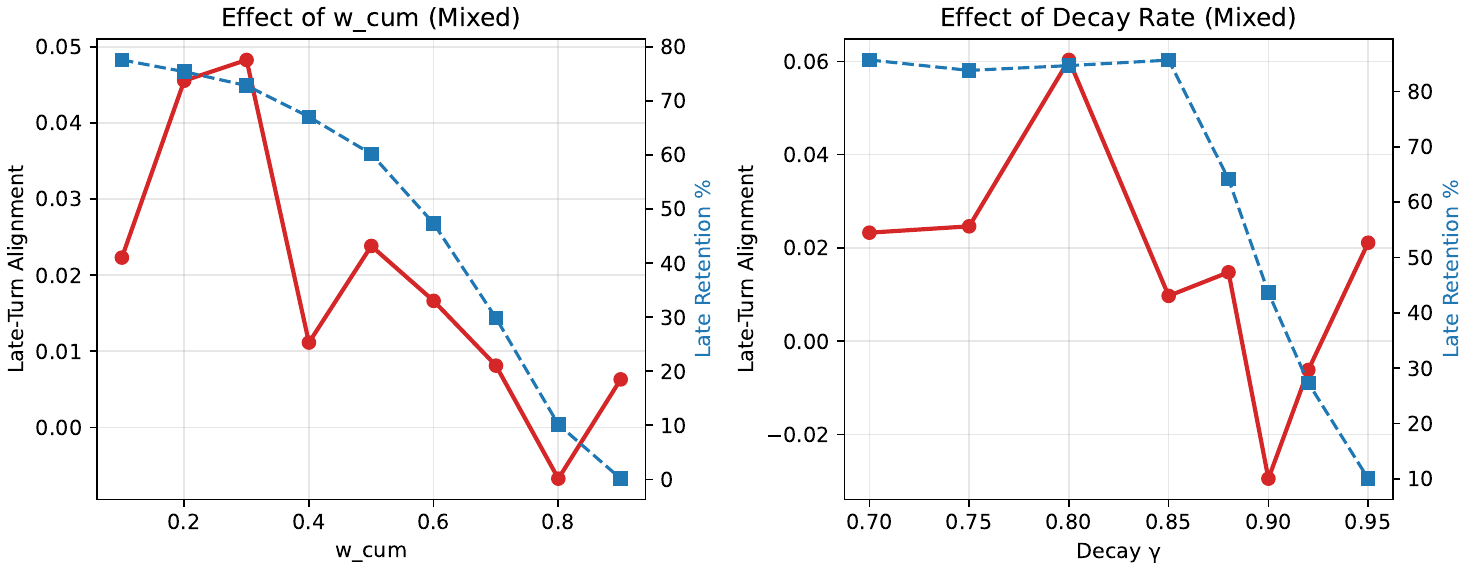}
    \caption{Ablation curves for $w_c$ (left) and $\gamma$ (right). Red: alignment; blue dashed: retention.}
    \label{fig:ablation}
\end{figure}

Figure~\ref{fig:ablation} visualizes both sweeps, showing the alignment--retention tradeoff. The ablation confirms that both channels contribute: removing either (extreme $w_c$ values) degrades performance.

\section{Discussion}
\label{sec:discussion}

\paragraph{Why Dual-Channel Scoring Works.}
The core insight of FD-KVC is disentangling \emph{long-term importance} (cumulative channel) from \emph{recent relevance} (recency channel). H2O's cumulative attention score monotonically increases and cannot decrease, causing the stale cache problem (Table~\ref{tab:adaptation}). FD-KVC's recency channel decays old relevance, enabling 3.6$\times$ faster adaptation, while the cumulative channel provides a stable floor that prevents premature eviction---explaining FD-KVC's superior diversity (80.6\% vs H2O's 70.0\%).

\paragraph{Reinforcement Learning Connection.}
The recency update (Eq.~\ref{eq:reinforce}) can be interpreted as a policy gradient where relevance $r_i$ serves as the reward signal \cite{williams1992simple}, analogous to REINFORCE. The adaptive learning rate (Eq.~\ref{eq:lr_adapt}) acts as a variance reduction mechanism, providing principled grounding distinct from ad-hoc scoring heuristics.

\paragraph{Practical Deployment.}
FD-KVC runs entirely on CPU (0.105\,ms/turn) as a drop-in cache manager requiring no model retraining.

\section{Conclusion}
\label{sec:conclusion}

We presented Fractional Decay KV-Cache (FD-KVC), a dual-channel scoring algorithm for memory-aware inference in dialog systems. By combining cumulative attention tracking with temporally decaying recency relevance, FD-KVC addresses the stale cache problem inherent in existing heavy-hitter methods. Across five multi-turn benchmarks, FD-KVC outperforms H2O by +6.7\% on composite alignment, adapts 3.6$\times$ faster to topic shifts, and achieves the highest topic diversity (80.6\%). The algorithm runs on CPU with negligible overhead and requires no model retraining.

Future work will extend FD-KVC to multi-head attention, evaluate on real-world dialog tasks (MultiWOZ, SGD), and explore integration with retrieval-augmented generation pipelines where ownership scores could inform cache priority for retrieved passages.

\section{Limitations}
\label{sec:limitations}

\paragraph{Synthetic Evaluation.}
All experiments use randomly projected 64-dimensional embeddings rather than real transformer representations. While this enables controlled comparison, it introduces variance that limits statistical significance at individual benchmark level. Validation with real LLM embeddings (e.g., LLaMA, GPT-family) is needed to confirm the improvements transfer to production settings.

\paragraph{Adaptation--Persistence Tradeoff.}
Temporal decay on the recency channel inherently trades long-term persistence for adaptation speed. On Topic Return, FD-KVC underperforms H2O by 39\% because decayed tokens cannot be recovered when the topic recurs. Applications requiring recall of distant context may need higher $w_c$ or an auxiliary retrieval mechanism.

\paragraph{Single-Head Evaluation.}
The current implementation evaluates single-head attention. In multi-head architectures, different heads attend to different aspects of context; FD-KVC's scoring would need to operate per-head or use a shared strategy, and the optimal design remains an open question.

\paragraph{Hyperparameter Sensitivity.}
FD-KVC introduces seven hyperparameters. While the ablation (Section~\ref{sec:ablation}) shows stable performance across a range, the optimal configuration may be task-dependent. Automated tuning would benefit deployment.

\paragraph{Computational Overhead.}
FD-KVC's latency (0.105\,ms/turn) is 3.9$\times$ FIFO due to dual-channel scoring and hybrid eviction. Although negligible relative to LLM decoding, it may become significant when applied at every transformer layer simultaneously.

\paragraph{Proxy Metrics.}
Our metrics (alignment, retention, diversity) are proxies for downstream task performance. Evaluation on end-to-end dialog tasks (response quality, slot filling, user satisfaction) would strengthen the conclusions.

\paragraph{Static Channel Weights.}
The weights $w_c$ and $w_\rho$ are fixed throughout a dialog. An adaptive mechanism that shifts toward recency during rapid topic changes and toward cumulative importance during stable phases could yield further improvements.

\section*{Acknowledgments}

The author thanks the SIGDIAL reviewers for their feedback on the submission and acknowledges the ACLPUB style-file maintainers for the camera-ready formatting resources used to prepare this version.


\appendix
\section{Hyperparameter Sensitivity}
\label{app:hyperparams}

Table~\ref{tab:hyperparams} provides the default hyperparameter values and their ranges explored during development.

\begin{table}[h]
\centering
\small
\begin{tabular}{@{}llll@{}}
\toprule
\textbf{Param} & \textbf{Default} & \textbf{Range} & \textbf{Description} \\
\midrule
$\gamma$ & 0.88 & [0.70, 0.95] & Recency decay rate \\
$\alpha_0$ & 0.30 & [0.10, 0.50] & Initial learning rate \\
$\tau$ & 0.02 & [0.01, 0.10] & Eviction threshold \\
$\beta$ & 0.25 & [0.10, 1.00] & Modulation exponent \\
$\mu$ & 0.40 & [0.10, 1.00] & Convergence factor \\
$w_c$ & 0.45 & [0.10, 0.90] & Cumulative weight \\
$w_\rho$ & 0.55 & [0.10, 0.90] & Recency weight \\
\bottomrule
\end{tabular}
\caption{FD-KVC hyperparameter configuration.}
\label{tab:hyperparams}
\end{table}

\section{Implementation Details}
\label{app:implementation}

All experiments were conducted on a single CPU core (Apple M-series, macOS). The implementation uses NumPy 1.24+ for vectorized operations. Each benchmark (600 dialogs $\times$ 4 methods) completes in under 60 seconds, confirming CPU efficiency. The complete source code is provided as supplementary material.

Key implementation choices:
\begin{itemize}
    \item Cumulative and recency scores are stored as two 1D float32 arrays, adding $8|\mathcal{C}|$ bytes of overhead.
    \item Original embeddings are stored alongside projected keys/values to compute embedding-space relevance.
    \item Cosine similarity for relevance computation is vectorized using batch matrix operations.
    \item Eviction uses \texttt{np.argsort} on the hybrid score for efficient top-$B$ selection.
\end{itemize}


\begin{thebibliography}{16}
\providecommand{\natexlab}[1]{#1}

\bibitem[{Ainslie et~al.(2023)Ainslie, Lee-Thorp, de~Jong, Zemlyanskiy,
  Lebr{\'o}n, and Sanghai}]{ainslie2023gqa}
Joshua Ainslie, James Lee-Thorp, Michiel de~Jong, Yury Zemlyanskiy, Federico
  Lebr{\'o}n, and Sumit Sanghai. 2023.
\newblock {GQA}: Training generalized multi-query transformer models from
  multi-head checkpoints.
\newblock In \emph{Proceedings of the 2023 Conference on Empirical Methods in
  Natural Language Processing (EMNLP)}.

\bibitem[{Bertsch et~al.(2023)Bertsch, Alon, Neubig, and
  Gormley}]{bertsch2024unlimiformer}
Amanda Bertsch, Uri Alon, Graham Neubig, and Matthew~R Gormley. 2023.
\newblock Unlimiformer: Long-range transformers with unlimited length input.
\newblock In \emph{Advances in Neural Information Processing Systems
  (NeurIPS)}.

\bibitem[{Bulatov et~al.(2023)Bulatov, Kuratov, and
  Burtsev}]{bulatov2024recurrent}
Aydar Bulatov, Yuri Kuratov, and Mikhail~S Burtsev. 2023.
\newblock Scaling transformer to 1m tokens and beyond with {RMT}.
\newblock In \emph{arXiv preprint arXiv:2304.11062}.

\bibitem[{Dao et~al.(2022)Dao, Fu, Ermon, Rudra, and
  R{\'e}}]{dao2022flashattention}
Tri Dao, Daniel~Y Fu, Stefano Ermon, Atri Rudra, and Christopher R{\'e}. 2022.
\newblock {FlashAttention}: Fast and memory-efficient exact attention with
  {IO}-awareness.
\newblock In \emph{Advances in Neural Information Processing Systems
  (NeurIPS)}.

\bibitem[{Ge et~al.(2024)Ge, Zhang, Liu, Zhang, Han, and Gao}]{ge2024model}
Suyu Ge, Yunan Zhang, Liyuan Liu, Minjia Zhang, Jiawei Han, and Jianfeng Gao.
  2024.
\newblock Model tells you what to discard: Adaptive {KV} cache compression for
  {LLMs}.
\newblock In \emph{International Conference on Learning Representations
  (ICLR)}.

\bibitem[{Kwon et~al.(2023)Kwon, Li, Zhuang, Sheng, Zheng, Yu, Gonzalez, Zhang,
  and Stoica}]{kwon2023efficient}
Woosuk Kwon, Zhuohan Li, Siyuan Zhuang, Ying Sheng, Lianmin Zheng, Cody~Hao Yu,
  Joseph~E Gonzalez, Hao Zhang, and Ion Stoica. 2023.
\newblock Efficient memory management for large language model serving with
  {PagedAttention}.
\newblock \emph{Proceedings of the ACM SIGOPS 29th Symposium on Operating
  Systems Principles}.

\bibitem[{Li et~al.(2024)Li, Huang, Yang, Vber, Hashimoto, Xiao, and
  Chen}]{li2024snapkv}
Yuhong Li, Yingbing Huang, Bowen Yang, Bharat Vber, Tatsunori Hashimoto,
  Guangxuan Xiao, and Beidi Chen. 2024.
\newblock {SnapKV}: {LLM} knows what you are looking for before generation.
\newblock \emph{arXiv preprint arXiv:2404.14469}.

\bibitem[{Liu et~al.(2023)Liu, Desai, Liao, Wang, Xie, Xu, Kyrillidis, and
  Shrivastava}]{liu2024scissorhands}
Zichang Liu, Aashiq Desai, Fangshuo Liao, Weitao Wang, Victor Xie, Zhaozhuo Xu,
  Anastasios Kyrillidis, and Anshumali Shrivastava. 2023.
\newblock Scissorhands: Exploiting the persistence of importance hypothesis for
  {LLM} {KV} cache compression at test time.
\newblock In \emph{Advances in Neural Information Processing Systems
  (NeurIPS)}.

\bibitem[{Mu et~al.(2023)Mu, Li, and Goodman}]{mu2024learning}
Jesse Mu, Xiang~Lorraine Li, and Noah~D Goodman. 2023.
\newblock Learning to compress prompts with gist tokens.
\newblock In \emph{Advances in Neural Information Processing Systems
  (NeurIPS)}.

\bibitem[{Pope et~al.(2023)Pope, Douglas, Chowdhery, Devlin, Bradbury, Heek,
  Xiao, Agrawal, and Dean}]{pope2023efficiently}
Reiner Pope, Sholto Douglas, Aakanksha Chowdhery, Jacob Devlin, James Bradbury,
  Jonathan Heek, Kefan Xiao, Shivani Agrawal, and Jeff Dean. 2023.
\newblock Efficiently scaling transformer inference.
\newblock \emph{Proceedings of Machine Learning and Systems}.

\bibitem[{Reid et~al.(2024)Reid, Savinov, Teber et~al.}]{devlin2024gemini}
Machel Reid, Nikolay Savinov, Denis Teber, and 1 others. 2024.
\newblock Gemini 1.5: Unlocking multimodal understanding across millions of
  tokens of context.
\newblock \emph{arXiv preprint arXiv:2403.05530}.

\bibitem[{Shazeer(2019)}]{shazeer2019fast}
Noam Shazeer. 2019.
\newblock Fast transformer decoding: One write-head is all you need.
\newblock In \emph{arXiv preprint arXiv:1911.02150}.

\bibitem[{Williams(1992)}]{williams1992simple}
Ronald~J Williams. 1992.
\newblock Simple statistical gradient-following algorithms for connectionist
  reinforcement learning.
\newblock In \emph{Machine Learning}, volume~8, pages 229--256.

\bibitem[{Wu et~al.(2024)Wu, Wang, Xiao, Peng, and Fu}]{wu2024retrieval}
Wenhao Wu, Yizhong Wang, Guangxuan Xiao, Hao Peng, and Yao Fu. 2024.
\newblock Retrieval head mechanistically explains long-context factuality.
\newblock \emph{arXiv preprint arXiv:2404.15574}.

\bibitem[{Xiao et~al.(2024)Xiao, Tian, Chen, Han, and
  Lewis}]{xiao2024efficient}
Guangxuan Xiao, Yuandong Tian, Beidi Chen, Song Han, and Mike Lewis. 2024.
\newblock Efficient streaming language models with attention sinks.
\newblock In \emph{International Conference on Learning Representations
  (ICLR)}.

\bibitem[{Zhang et~al.(2023)Zhang, Sheng, Zhou, Chen, Zheng, Cai, Song, Tian,
  R{\'e}, Barrett, Wang, and Chen}]{zhang2024h2o}
Zhenyu Zhang, Ying Sheng, Tianyi Zhou, Tianlong Chen, Lianmin Zheng, Ruisi Cai,
  Zhao Song, Yuandong Tian, Christopher R{\'e}, Clark Barrett, Zhangyang Wang,
  and Beidi Chen. 2023.
\newblock {H2O}: Heavy-hitter oracle for efficient generative inference of
  large language models.
\newblock In \emph{Advances in Neural Information Processing Systems
  (NeurIPS)}.

\end{thebibliography}
\end{document}